\documentclass{article}
\usepackage{stywhispers,amsmath,epsfig}

\usepackage{amsfonts,amssymb}
\usepackage{overpic}
\usepackage{graphicx}
\graphicspath{ {./img/} }
\usepackage[caption=false]{subfig}
\usepackage[colorlinks=true, allcolors=blue]{hyperref}
\usepackage[ruled,vlined,linesnumbered,noresetcount]{algorithm2e}
\usepackage{amsmath}
\usepackage{multicol}
\usepackage{multirow}
\usepackage{hyperref}

\title{STARS: Sensor-agnostic Transformer Architecture for Remote Sensing}
\name{Ethan King, Jaime Rodriguez, Diego Llanes, Timothy Doster, Tegan Emerson, James Koch
\thanks{This work was conducted under the Laboratory Directed Research and Development Program at PNNL, a multi-program national laboratory operated by Battelle for the U.S. Department of Energy under contract DE-AC05-76RL01830. Contact: \href{malito:james.koch@pnnl.gov}{james.koch@pnnl.gov}. 
 }}
\address{Pacific Northwest National Laboratory \\ Richland, WA}

\begin{document}
%
\maketitle
\begin{abstract}
We present a sensor-agnostic spectral transformer as the basis for spectral foundation models. To that end, we introduce a Universal Spectral Representation (USR) that leverages sensor meta-data, such as sensing kernel specifications and sensing wavelengths, to encode spectra obtained from any spectral instrument into a common representation, such that a single model can ingest data from any sensor. Furthermore, we develop a methodology for pre-training such models in a self-supervised manner using a novel random sensor-augmentation and reconstruction pipeline to learn spectral features independent of the sensing paradigm. We demonstrate that our architecture can  learn sensor independent spectral features that generalize effectively to sensors not seen during training. This work sets the stage for training foundation models that can both leverage and be effective for the growing diversity of spectral data.
\end{abstract}
\section{Introduction}
\label{sec:intro}
The proliferation of Earth Observation (EO) instruments has led to an explosion in the amount of data derived from sensing platforms, each possessing a unique spectral response in support of different scientific missions \cite{sudmanns2020big}. Paired with this proliferation of sensors and data is a corpus of Machine Learning (ML) and statistics-based exploitation tools \cite{salcedo2020machine}. These tools are typically crafted to leverage a specific data modality for a specific task; e.g. using Landsat-derived data for quantifying vegetation health with a bespoke ML model (e.g. \cite{burchfield2016application, shamsoddini2018mapping, balogun2020spatio}). The pace of proliferation in these spaces is increasing \cite{belward2015launched}, which poses unique challenges. First, there is technical risk and unnecessary overhead associated with a one-cube, one-task, one-model exploitation pipeline. Second, leveraging multiple modalities for a particular task is not possible (without some form of interpolation), which is challenging when analysis pipelines may need to accommodate past, current, and future sensing platforms. 

The Foundation Modeling paradigm has emerged to address these challenges: Foundation Models (FMs) seek to leverage exposure to large amounts of unlabelled diverse data, such that relevant features and relationships may be extracted and leveraged for downstream tasks \cite{bommasani2021opportunities}. FMs have enjoyed great successes in Natural Language Processing (NLP), though their use in domain-specific contexts is still limited. For spectral modalities, FMs have typically inherited the constraint of a single sensor data source; for example, fixing the model architecture to expect specific number of bands in a particular order, but allowing a downstream task to change. 

SpectralGPT \cite{hong2024spectralgpt} is one such example of a FM that leverages large-scale data from a single sensor that can be adapted to many downstream tasks, such as classification, segmentation, and change detection. SpectralGPT is built upon a spectral-spatial vision transformer and is trained in a self-supervised manner with masked autoencoding. Similarly, SpectralFormer \cite{hong2021spectralformer} was proposed as a backbone network architecture for extracting high-quality representations from spectral data by first performing a group-wise spectral embedding; i.e. capturing spectrally-local continuous spectral information. A similar strategy is adopted in HyperSIGMA \cite{wang2024hypersigma}, a large foundation model with over 1 billion parameters, which performs a local spectral clustering into a desired feature dimension to accommodate varying number of bands. SS-MAE \cite{lin2023ss} is a FM that emphasized multi-modal encoding for joint classification tasks. SS-MAE leverages a similar transformer/CNN architecture to encode these modalities into a common representation. Lastly, Dynamic-One-For-All (DOFA) \cite{xiong2024neural} is a recently proposed FM that presents `neural plasticity,' or the ability to adapt weights based on input metadata. Here, wavelength is a unifying parameter across all modalities. 

Despite recent advances in FMs in this domain, several issues persist. Many models assume specific input modalities, typically relying on data from specific instruments (e.g. Landsat, Sentinel, AVIRIS, etc.), which precludes the ability to generalize to different sensors and environments. Furthermore, many models treat spectral data as sequences of measurements at specific wavelengths, utilizing convolutions, pooling layers, or group-wise embeddings for feature extraction. However, spectral measurements are not inherently sequential, but rather a ``grab-bag'' of unique measurements which may or may not be correlated across wavelengths. This poses an additional challenge: current modeling approaches cannot effectively distinguish between shuffled bands, recognize redundant information, nor discern between measurements obtained from narrow (e.g. hyperspectral) or wide (e.g. RGB) bands. 

In this work, we narrow our focus specifically to enabling sensor-agnostic FMs with robust and consistent treatment of spectral signatures. Towards this end, this work presents \textit{Sensor-Agnostic Spectral Transformers}, an architecture for physically-consistent feature extraction and encoding for sensor-agnostic spectral inputs. Our core contributions include: (i) definition of a \textit{Universal Spectral Representation}, (ii) a resolution-independent spectral emulator, and (iii) a framework for training the proposed models by using novel sensor data agumentations. Our approach aims to establish a universal, sensor-agnostic framework for spectral science, enabling the development of more robust and versatile spectral FMs. We demonstrate our architecture's utility towards these tasks by comparing latent representations of spectral signatures derived from unseen sensors, showing that our methods can faithfully extract pertinent information across sensing paradigms. 

\begin{figure*}[]
\centering
        \begin{overpic}[width=\linewidth]{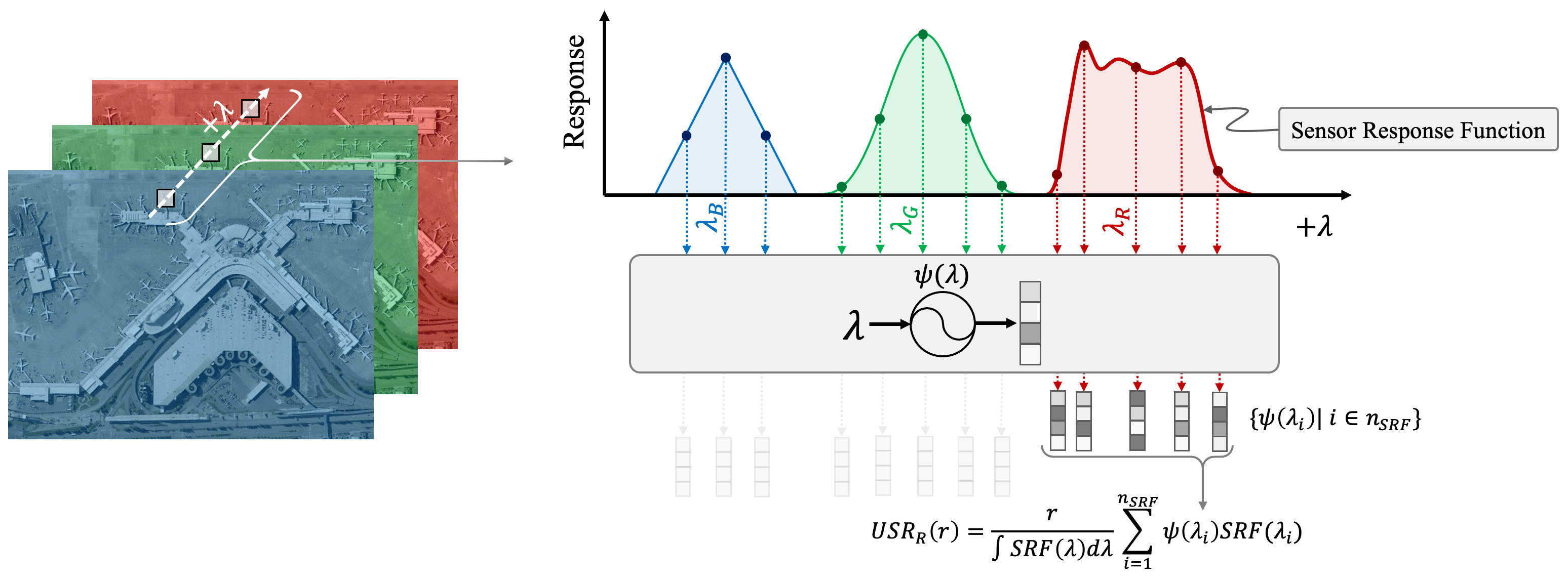}
        \put(1,35){(a)}
        \put(33,35){(b)}
        \end{overpic}  
        \caption{A multi-band image is displayed in (a) where each channel corresponds to a different band of a spectral sensor. In (b), the notional Sensor Response Function (SRF) for each sensor band is depicted. A pixel from the image in (a) contains a measurement value corresponding to each SRF; in this case, each pixel is an RGB triplet. These measurements contain no information about the originating SRFs, which presents a challenge: without special treatment, a model cannot distinguish between shuffled bands, recognize redundant information, or discern between measurements obtained from narrow (hyperspectral) or wide (multi-band) SRFs. Towards this end, we introduce \textit{Universal Spectral Representations} (USRs, shown in (b)), which encode spectral measurements into a common representation that contains both measurement values as well as SRF and spectral coverage information. USRs act as an approximation for real sensor physics, leveraging positional encodings and known SRFs to produce a feature-rich representation of band-wise measurements that are consistent for any SRF.}
        \label{fig:usr}
\end{figure*}

\section{Methodology}
\label{sec:methods}

\subsection{Architecture Overview}
Our proposed architecture contain three distinct modules: (i) a novel tokenization scheme that we term \textit{Universal Spectral Representations} (USRs), (ii) a spectral transformer encoder, and (iii) an operator-theoretic decoder. We pair these methods with a robust data augmentation strategy to achieve sensor-agnostic model training. 

A spectral signature is passed to the model as a set of measurements and a description of the corresponding band Spectral Response Functions (SRF). Each measurement is mapped to a vector in $\mathbb{R}^{n}$ by the USR as defined in Section \ref{sec:USR}. A standard transformer architecture is then used as an encoder to handle this unordered and variable number of vector inputs. The transformer output is then reduced to a target embedding dimension by a linear layer. 

To demonstrate our sensor-agnostic spectral transformer, we present a self-supervised reconstruction task to reconstruct a high resolution spectra from a lower resolution sensor view of the same scene over a large distribution of random sensor augmentations. For reconstruction, the embedding is passed to a Deep Operator Network (DeepONet) as detailed in \ref{sec:DON}.

\subsection{Universal Spectral Representation}
\label{sec:USR}
A sensor may be comprised of any number of diverse bands of varying shape. Our goal is to construct a flexible representation that can easily handle these diverse inputs. As depicted in Figure \ref{fig:usr}, ideally we could directly feed a model the SRF of each band to provide additional context for a measured value, but this would of course be cumbersome. Instead we propose a method to quickly approximate the SRF as a linear function. 

We make the assumption that the relevant information for model input is how measured radiance for a band may be appropriately distributed over wavelengths. For a measured value $r$ and SRF, we would like to input $ r \text{SRF}(\lambda)$.

To start we borrow the sinusoid positional encoding frequently employed in language transformers as a means of encoding wavelength as an $n$ dimensional vector. For a wavelength $\lambda \in [0, \infty)$ let $\psi (
\lambda)$ be the mapping $\mathbb{R} \to \mathbb{R}^{n}$ defined by 
\begin{equation}
\label{eq:sinusoid_def}
\psi(\lambda)_{i}= \begin{cases} sin(\lambda\sigma^{\frac{2i}{n}} \frac{n}{r}) & \mbox{ i even} \\
cos(\lambda\sigma^{\frac{2(i-1)}{n}} \frac{n}{r}) & \text{ i odd}
\end{cases} \, ,
\end{equation}
where $\sigma$ and $r$ are hyperparameters that can be used to tune the encoding. 

We then seek to construct a mapping $\text{USR}: \mathbb{R} \to \mathbb{R}^{n}$ for the given SRF such that
\begin{equation}
\langle \text{USR}(r), \psi(\lambda)\rangle ~ \approx ~ r \text{SRF}(\lambda) \, ,
\label{eq:USR_band_approx}
\end{equation}
where $\langle \cdot,\cdot \rangle$ denotes the inner product. The output vector $\text{USR}(r)$ will serve as our representation of the band and measurement.

Let $\{ \lambda_{i} \}_{i=1}^{n_{\text{SRF}}}$ a set of $n_{\text{SRF}}$ wavelengths spanning the nonzero region of the SRF, we then set
\begin{equation}
\text{USR}(r) = \frac{r}{\int \text{SRF}(\lambda)d\lambda}\sum_{i=1}^{n_{\text{SRF}}} \psi(\lambda_{i})\text{SRF}(\lambda_{i}) \,.
\label{eq:usr definition}
\end{equation}

\begin{figure}[ht]
\label{fig:band_approximation}
\includegraphics[width=8cm]{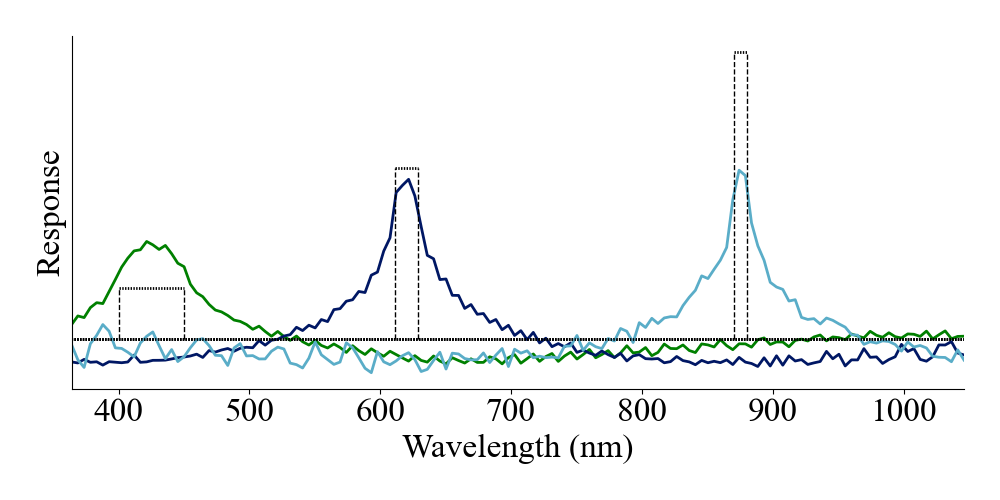}
\caption{Three notional Sensor Response Functions (SRF) are shown as dashed rectangles of different widths and heights (as scaled by sensor input). Through recasting these SRFs into our Universal Spectral Representation (USR, Eq. \ref{eq:USR_band_approx}), the combined measurement-SRF can be fed into our transformer architecture. Critically, our choice for constructing a USR allows one to approximately recover the embedded information, such as band width and height, as shown here as solid lines.}
\end{figure}

Figure \ref{fig:band_approximation} shows an example for the hyperparameter choices used in this paper for how the USR as defined in \eqref{eq:usr definition} can approximate band SRFs as given in the expression \eqref{eq:USR_band_approx}. The approximation is quite crude and for example the USR approximation does not scale in direct proportion to the magnitude of the measured radiance. However, as we report in Section \ref{sec:results} it can be sufficient to convey generalizable sensor information. It is important to note that the behavior of this representation is very sensitive to the hyperparameter choices for the sinusoid encoding, and the results presented here only cover one such case.

\subsection{Operator-Theoretic Decoder}
\label{sec:DON}

To remain sensor-agnostic, our reconstruction task necessitates a decoder capable of evaluating at any resolution, ensuring the architecture is not constrained to a specific sensor modality. We achieve this by employing a Deep Operator Network (DeepONet) \cite{lu2019deeponet}, which approximates spectral signatures at any wavelength based on the latent representation of the model input. We implement the specific architecture proposed in \cite{Sifan2022deeponet}.

\subsection{Data Augmentation}
\begin{figure}[ht]
\includegraphics[scale = 0.25]{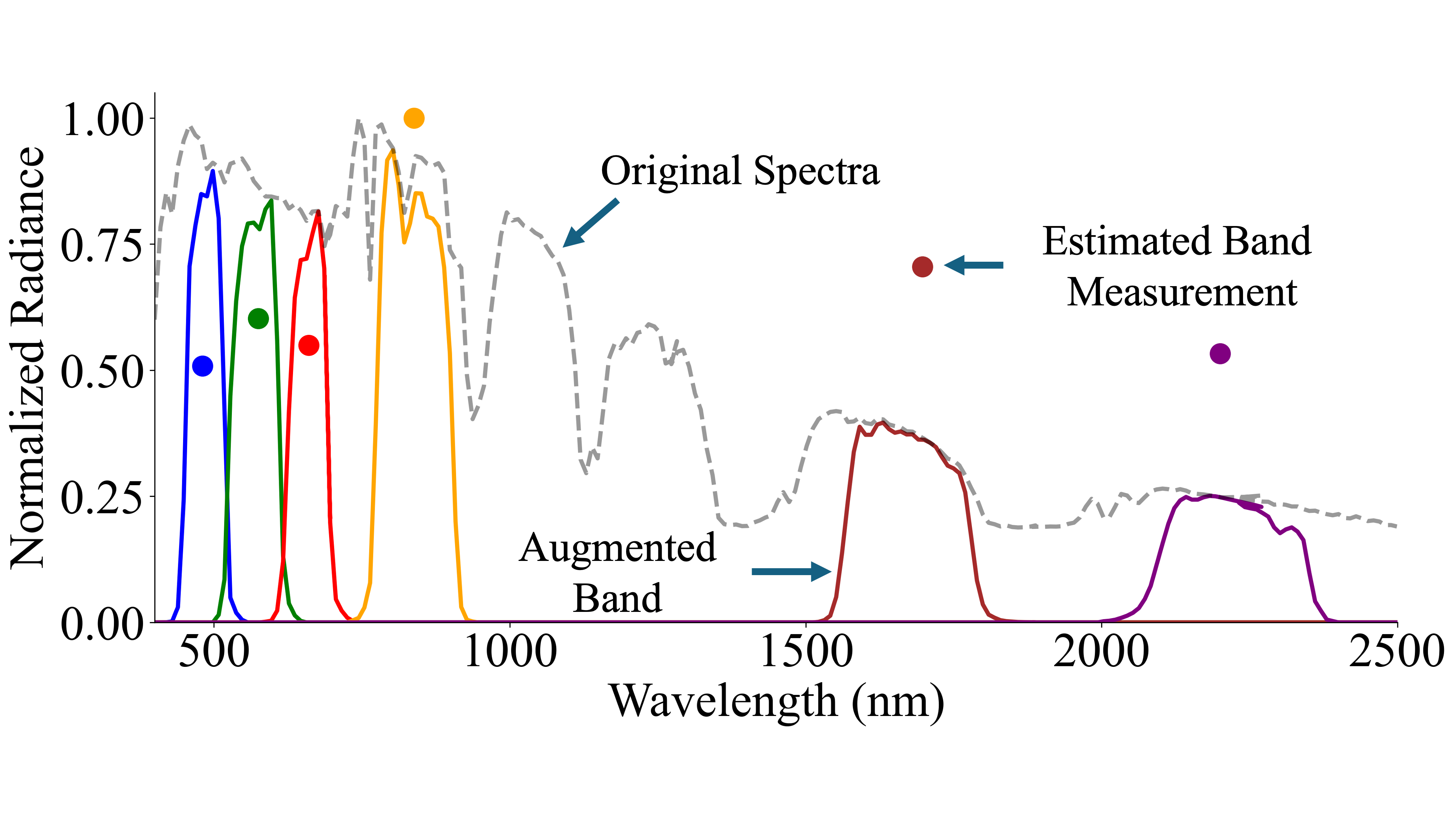}
\caption{Graphical representation of the augmentation module on single-pixel spectra. The product of  SRF bands with the original spectra produces the augmented bands, and integrating those products produces an estimated measurement for each band.}
\label{fig:spectral_convolution}
\end{figure}
The scarcity of spectral data representing diverse environmental conditions, spectral characteristics, and sensor configurations imposes limitations to the development of robust spectral foundation models. We address these limitations by augmenting existing spectral data, thereby creating new and diverse scenes which are independent of the resolution and instrument from which the original spectra were collected.

The sensor augmentation module resamples spectra to different resolutions via spectral convolution, as shown in Eq.(4). Here,  \(\lambda_{\text{min}}\) and  \(\lambda_{\text{max}}\) represent the wavelength range. The SRF for the \( i \)-th band of a sensor is \(\text{SRF}_i(\lambda) \), where \( i = 1, 2, \ldots, N \) and \( N \) is the number of bands in the sensor, while the input spectral radiance for a scene is represented by \(x(\lambda)\). The spectral convolution \( r_i \) for the \( i \)-th band is then defined as the integral of the product of \( x(\lambda)\)  and \( \text{SRF}_i(\lambda)\), normalized by the integral of the SRF. 
\begin{equation}
r_i = \frac{\int_{\lambda_{\text{min}}}^{\lambda_{\text{max}}} x(\lambda) \text{SRF}_i(\lambda)d\lambda}{\int_{\lambda_{\text{min}}}^{\lambda_{\text{max}}} \text{SRF}_i(\lambda)d\lambda} .
\label{eq:convolution}
\end{equation}
This convolution is repeated for each band in the SRF, thereby providing an approximation of a new view of the original scene from a different sensor. A graphical representation of this augmentation process is depicted in Figure \ref{fig:spectral_convolution}.

\subsection{Model Training}
We train the model in a sensor-agnostic manner by minimization of reconstruction loss over a large distribution of sensor augmentations. Let $\mathcal{S}$ be a set of sensor augmentations. We train the model comprised of the encoder $e$ and decoder $d$, to minimize the loss
\begin{equation}
\mathcal{L} = \sum_{ (x,s) \in \mathcal{X} \times \mathcal{S}} M(x, d(e(s(x))) )  \, ,
\end{equation}
where $\mathcal{X}$ is a set of spectral training data, and $M$ is an appropriate reconstruction metric. Here we performed in-scene training using cosine similarity for a metric, full details are given in Section \ref{sec:experiment}

\section{Results}
\label{sec:results}

\subsection {Experimental Setup}
\label{sec:experiment}
All model training was conducted on the Houston hyperspectral dataset \cite{HoustonData}. The data is comprised of a hypersepctral image of the University of Houston Campus from a Compact Airborne Spectographic Imager (CASI) with 144 spectral bands convering 380 nm to 1050 nm at a roughly 5 nm resolution. Almost 3000 pixels are labeled into 15 classes of equal sizes. Test and train sets were constructed by random 50/50 class balanced splitting such that approximately 200 pixels of each class are represented in both test and train splits. Experiments were carried out on five such splits for cross validation.

The sinusoid wavelength encoding \eqref{eq:sinusoid_def} was computed with $\{\sigma = 3\text{e-}4 , r = 10 , n=200 \}$. For the USR \eqref{eq:usr definition} we assume sensor SRFs are square waves with width equal to the sensor full width half maximum, and we sample wavelenghts with resolution 1nm. Both the transformer encoder and DeepONet decoder had 3 layers with hidden dimension 250 and 200 respectively. The transformer had 10 heads and learned a 3 dimensional embedding of the data. 

For data augmentation random sensors were generated at batch time with non-overlapping square kernels for bands. The number of bands for each sensor was sampled randomly from between 3 and 144 with a truncated exponential distribution using decay rate $0.1$. Center locations of bands were sampled uniformly from $350nm$ to $1050nm$ with step size $5nm$, and band width was sampled uniformly from $1nm$ up to $200nm$ or the maximum size to prevent overlapping neighboring bands.

The model was trained for 1500 epochs using the Adam optimizer with a reduce on plateau learning rate scheduler starting at $1\text{e-}3$. Model training was done using an A100 GPU and took approximately 3 hours per experiment.
\subsection{Reconstruction from unseen sensor inputs}

To test the effectiveness of our sensor-agnostic methodology we evaluate our learned model using four unseen sensors; Landsat 5 TM (L5 TM), only the red, blue and green bands of L5 TM (RGB), the Advanced Land Imager (ALI), and CASI the sensor used to collect the data. Bands of sensors that fall outside of the range of the available data are ignored. We measure reconstruction performance with cosine similarity as it quantifies shape, which (in the case of spectroscopy) is widely used  for identification and characterization of materials. In particular for two spectra $x$ and $y$ we report $1 - \frac{\langle x, y \rangle}{||x|| ||y||}$ which will be zero if the spectra are identical up to scaling, and approach one as they become more dissimilar.

\begin{figure}[h!]
\includegraphics[width=8cm]{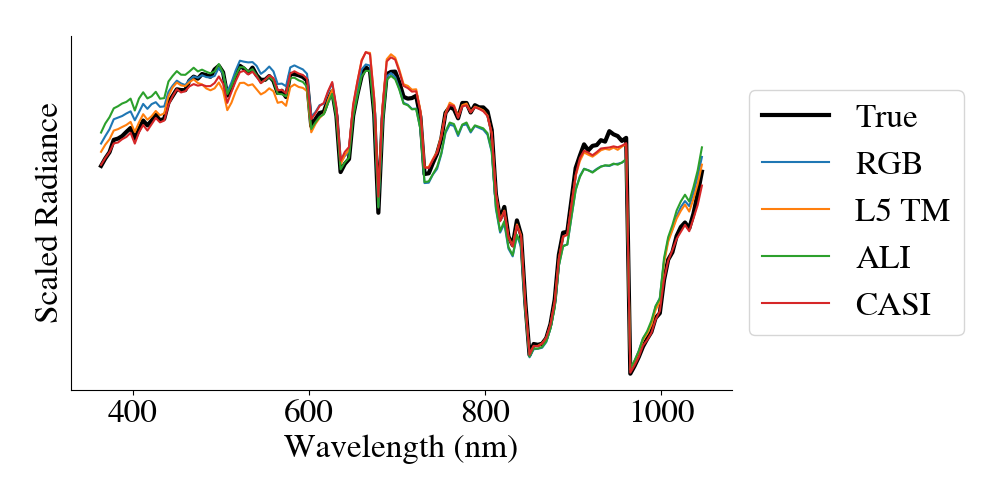}
\caption{Comparison of model spectral reconstruction on a test set pixel viewed through four different sensors not seen during training as inputs.}
\label{fig:reconstruction_example}
\end{figure}

As shown in Table \ref{table:recon_acc} and Figure \ref{fig:reconstruction_example} the model is able to produce close reconstructions using unseen sensor inputs. Moreover, the reconstruction improves as the sensor information is increased as demonstrated by the more than a third improvement in cosine similarity when using the CASI sensor inputs over the others. As a whole the real sensors also produce better reconstructions on average than random sensor inputs as shown in Table \ref{table:recon_acc}, agreeing with the intuition that sensors with bands chosen to capture more information will improve performance.

\begin{table}[!h]
\centering
\resizebox{\columnwidth}{!}{
\begin{tabular}{ p{0.18\linewidth}| p{0.12\linewidth} p{0.12\linewidth} p{0.12\linewidth} p{0.12\linewidth} p{0.12\linewidth} }
 Sensor & Random & RGB & L5 TM & ALI & CASI \\
 \hline
Bands & $\sim$ 5 & 3 & 4 & 7 & 144 \\
Cos sim & 0.013 & 0.0027 & 0.0027 & 0.0022 & 0.00069 \\
Stdev & 1.9E-3  & 2.7E-4 & 3.0E-4 & 9.7E-5 & 1.3E-4  
\end{tabular}
}
\caption{Average cosine similarity (Cos sim) and standard deviation (Stdev) is reported between the original and model reconstructed spectra on the test set over five cross validation trials on each of four sensors not seen during training. Random sensors were also drawn as is done during each training iteration for comparison, with on average 5 bands per sensor.}
\label{table:recon_acc}
\end{table}

\begin{figure}[h!]
\centering
        \begin{overpic}[width=\linewidth]{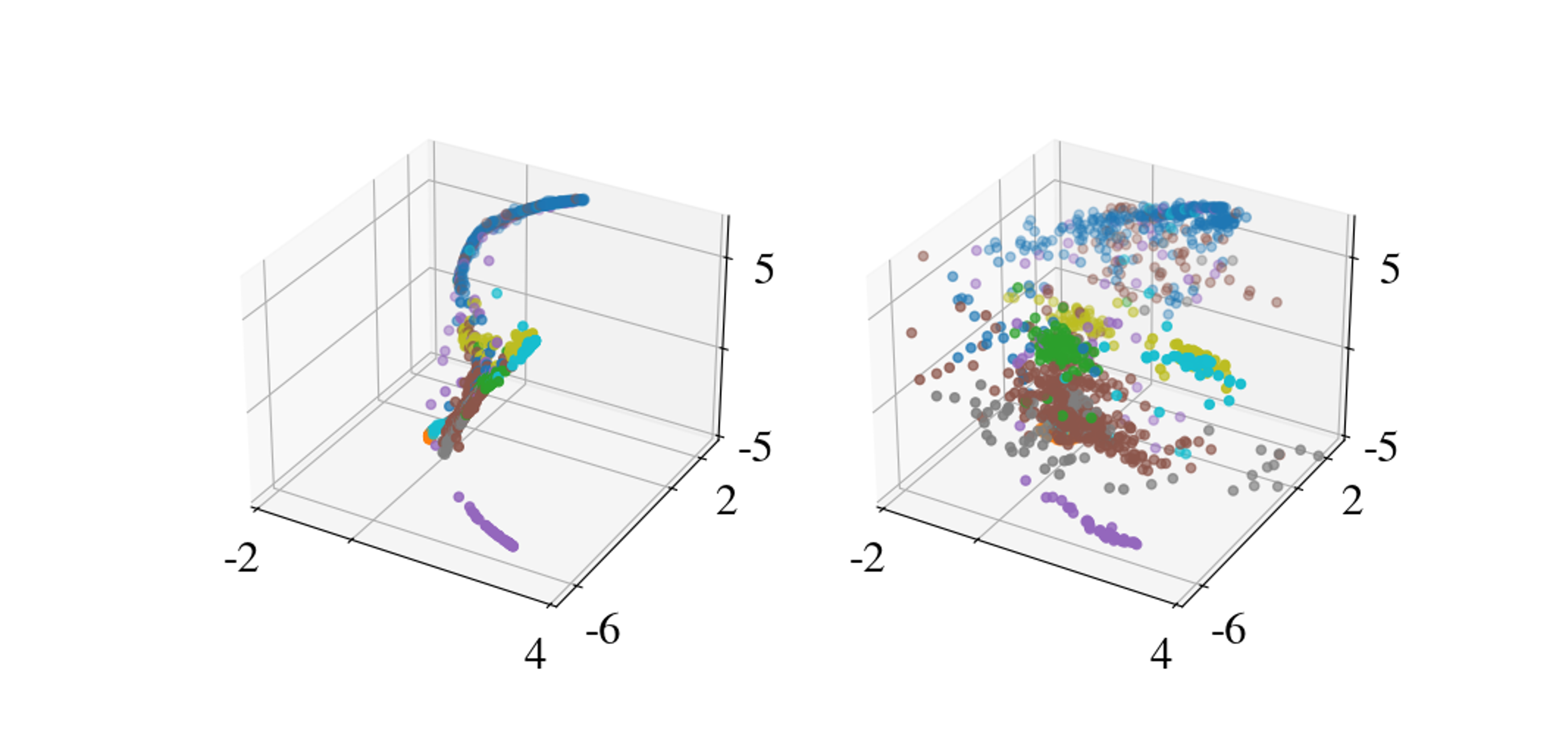}
        \put(14,35){(a)}
        \put(53,35){(b)}
        \end{overpic}  
        \caption{A visualization of the three dimensional learned model embedding for one of the experiments. Plots are the model encoder output from test set pixels input to the model through a) an RGB sensor, and b) a CASI sensor. The points are colored by the class label for the corresponding pixel. For clarity the Houston dataset classes (Healthy Grass, Stressed Grass, Trees) and (Road, Highway, Parking Lot 1, Parking Lot 2) were combined into two classes vegetation and asphalt respectively. Further, the two mixed pixel classes residential and commercial are excluded from the plot. }
        \label{fig:embedding_comparison}
\end{figure}

The compression of the data by the model encoder down to a three dimensional embedding makes for easy visualization of the learned representation as shown in Figure \ref{fig:embedding_comparison}. Across all the experiments we see a clear separation of the data in the embedding space by class, and further see that the test set embedding distribution remains relatively stable over sensor inputs as reported in Table \ref{table:emb_cmpr}. Over the wavelengths in the dataset, L5 TM has just one additional band beyond RGB, and their respective embeddings are found to be most close together. Similarly, ALI has relatively few bands in comparison to CASI and its embedding is found to be closer to that of L5 TM and RGB. The stability of the embedding over unseen sensor inputs is a promising result for the utility of the learned spectral representation by the model for downstream tasks in a sensor-agnositc manner.
\begin{table}[!h]
\centering
\resizebox{\columnwidth}{!}{
\begin{tabular}{ p{0.2\linewidth}| p{0.08\linewidth} p{0.08\linewidth} p{0.08\linewidth} p{0.08\linewidth} p{0.08\linewidth} p{0.08\linewidth} }
 Comparison & RGB  L5TM & RGB  ALI & RGB  CASI & L5TM ALI & L5TM  CASI & ALI CASI\\
 \hline
Relative Pixel Shift & \hspace{1cm}  0.12 &  \hspace{1cm} 0.18 & \hspace{1cm} 0.24 & \hspace{1cm} 0.15 & \hspace{1cm}  0.24 & \hspace{1cm} 0.22 \\
Stdev & 0.025 & 0.038 & 0.039 & 0.051 & 0.032 & 0.024
\end{tabular}
}
\caption{For each pair of sensors in the column heading, the table reports the average distance between the embedding of each test set pixel viewed through the two sensors divided by the average cross distance between all pixels. Reported values are the average and standard deviation (Stdev) over five cross validation trials.}
\label{table:emb_cmpr}
\end{table}

\section{Discussion and Conclusion}
\label{sec:discussion}
This work presents a first-of-its-kind ML architecture specifically tailored to accept spectral inputs from any sensor, enabling robust and consistent treatment of spectra for the next-generation of spectral foundation models. Our architecture has several unique properties. First, this architecture can handle shuffled and redundant inputs effectively by leveraging known Sensor Response Functions. Furthermore, this framework allows for the integration of imagery from different modalities, such as combining RGB data with more expensive hyperspectral data like AVIRIS. Our architecture's ``grab-bag" style of data handling, coupled with training on random sensor data, offers a future-proof solution that maintains independence of input order and robust performance across diverse sensor inputs.

As presented, our pretraining strategy focused exclusively on scene-specific pretraining with random sensor augmentations. While effective for our immediate goals, a more comprehensive foundation model will require an augmentation scheme based on an endmember library, sensor augmentation, and atmospheric transmission model, effectively creating a pseudo-infinite data set. This approach aims to better generalize across various sensors and conditions. Furthermore, recognize that for such a model to be fully effective, a thorough analysis and optimization of hyperparameters is essential. However, we have relegated this task to future work, acknowledging that it is a necessary step for maximizing the model's performance. Lastly, we have not yet explored connections to spatial information, such as in spectral-spatial models, in this study. Our current work assumes pure pixels without incorporating spatial context. We reserve spectral-spatial analysis for future research, recognizing that integrating spatial information is a crucial next step for advancing the capabilities of our model.

The development of a sensor-agnostic transformer architecture for spectral foundation models marks a significant advancement in the field of spectral sciences. By effectively integrating sensing kernel information and creating a Universal Spectral Representation (USR), our approach addresses the complexities and diversities of spectral data across various sensors. This architecture enhances model generalization and robustness but also enables more versatile applications in remote sensing and hyperspectral imaging. 

\bibliographystyle{IEEEbib}
\bibliography{main}

\end{document}